\documentclass{article}

\usepackage{microtype}
\usepackage{booktabs} 

\usepackage{hyperref}

\usepackage[accepted]{icml2019}

\usepackage{amsmath}
\usepackage{comment}
\usepackage{enumitem,kantlipsum}
\usepackage{collcell}
\usepackage{xcolor,colortbl}
\usepackage{pgfplots}
\usepackage{tikz}
\usepackage{amssymb}
\usepackage{graphicx}
\usepackage{subcaption}
\usepackage{multirow}
\usepackage{verbatimbox}

\usepackage[disable]{todonotes}

\newcommand{\etal}{\textit{et al}.}
\newcommand{\ie}{\textit{i}.\textit{e}.}
\newcommand{\eg}{\textit{e}.\textit{g}.}

\begin{document}

\twocolumn[
\icmltitle{Unsupervised Data Uncertainty Learning in Visual Retrieval Systems}



\icmlsetsymbol{equal}{*}

\begin{icmlauthorlist}
\icmlauthor{Ahmed Taha}{umd}
\icmlauthor{Yi-Ting Chen}{honda}
\icmlauthor{Teruhisa Misu}{honda}
\icmlauthor{Abhinav Shrivastava}{umd}
\icmlauthor{Larry Davis}{umd}
\end{icmlauthorlist}

\icmlaffiliation{umd}{Univeristy of Maryland, College Park}
\icmlaffiliation{honda}{Honda Research Institute, USA}

\icmlcorrespondingauthor{Ahmed Taha}{ahmdtaha@cs.umd.edu}

\icmlkeywords{Machine Learning, ICML}

\vskip 0.3in
]



\printAffiliationsAndNotice{}  

\begin{abstract}

	We introduce an unsupervised formulation to estimate heteroscedastic uncertainty in retrieval systems. We propose an extension to triplet loss that models data uncertainty for each input. Besides improving performance, our formulation models local noise in the embedding space. It quantifies input uncertainty and thus enhances interpretability of the system. This helps identify noisy observations in query and search databases. Evaluation on both image and video retrieval applications highlight the utility of our approach. We highlight our efficiency in modeling local noise using two real-world datasets: Clothing1M and Honda Driving datasets. Qualitative results illustrate our ability in identifying confusing scenarios in various domains. Uncertainty learning also enables data cleaning by detecting noisy training labels.

\end{abstract}

\section{Introduction}

\begin{figure}
	\centering
	\setlength{\fboxsep}{0pt}%
	\setlength{\fboxrule}{2pt}%
	
	\fcolorbox{green}{white}{\includegraphics[height=0.25\linewidth]{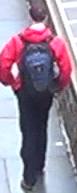}}
	\fcolorbox{green}{white}{\includegraphics[height=0.25\linewidth]{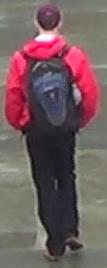}}
	\fcolorbox{green}{white}{\includegraphics[height=0.25\linewidth]{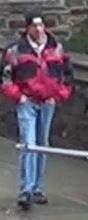}}
	\fcolorbox{green}{white}{\includegraphics[height=0.25\linewidth]{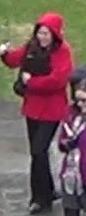}}
	\fcolorbox{green}{white}{\includegraphics[height=0.25\linewidth]{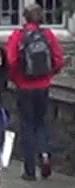}}

~

	\fcolorbox{red}{white}{\includegraphics[height=0.25\linewidth]{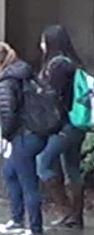}}
	\fcolorbox{red}{white}{\includegraphics[height=0.25\linewidth]{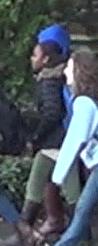}}
	\fcolorbox{red}{white}{\includegraphics[height=0.25\linewidth]{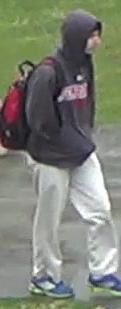}}
	\fcolorbox{red}{white}{\includegraphics[height=0.25\linewidth]{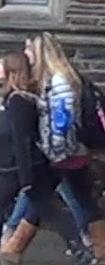}}
	\fcolorbox{red}{white}{\includegraphics[height=0.25\linewidth]{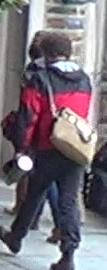}}
		\caption{The first and second rows show five lowest and highest  uncertainty queries (respectively) identified from DukeMTMC-ReID dataset.}
		\label{fig:duke_confusin_person}
\end{figure}

Noisy observations hinder learning from supervised datasets. Adding more labeled data does not eliminate this inherent source of uncertainty. For example, object boundaries and objects farther from the camera remain challenging in semantic segmentation, even for humans. Noisy observations take various forms in visual retrieval. 
The noise can be introduced by a variety of factors; e.g., low resolution inputs, a wrong training label.
Modeling uncertainty in training data can improve both the robustness and interpretability of a system. In this paper, we propose a formulation to capture data uncertainty in retrieval applications. Figure~\ref{fig:duke_confusin_person} shows the lowest and highest uncertainty query images, detected by our system, from DukeMTMC-ReID person re-identification dataset. Similarly, in autonomous navigation scenarios, our formulation can identify confusing scenarios; thus improving the retrieval efficiency and interpretability in this safety-critical system.

Labeled datasets contain observational noise that corrupts the target values~\cite{bishop1995neural}. This  noise, also known as aleatoric uncertainty ~\cite{kendall2017uncertainties}, is inherent in the data observations and cannot be reduced even if more data is collected. Aleatoric uncertainty is categorized into homoscedastic and heteroscedastic uncertainty. Homoscedastic uncertainty is task dependent, i.e., a constant observation noise $\sigma$ for all input points. On the contrary, heteroscedastic uncertainty posits the observation noise $\sigma(x)$ as dependent on input $x$. Aleatoric uncertainty has been modeled in regression and classification applications like per-pixel depth regression and semantic segmentation tasks respectively. In this paper, we extend triplet loss formulation to model heteroscedastic uncertainty in retrieval applications.

Triplet loss~\cite{schroff2015facenet} is a prominent ranking loss for space embedding. It has been successfully applied in face recognition~\cite{schroff2015facenet,sankaranarayanan2016triplet} and person re-identification~\cite{cheng2016person,su2016deep,ristani2018features}. In this paper, we extend it to capture heteroscedastic uncertainty in an unsupervised manner. Vanilla triplet loss assumes a constant uncertainty for all input values. By integrating the anchor, positive, and negative uncertainties $(\sigma_a,\sigma_p,\text{and } \sigma_n \text{ respectively})$ in the loss function, our model learns data uncertainty nonparametrically. Thus, the data uncertainty becomes a function of different inputs, i.e., every object has a different $\sigma(x)$.

We evaluate our unsupervised formulation on two image retrieval applications: person re-identification and fashion item retrieval. Person re-identification datasets provide an established quantitative evaluation benchmark. Yet, they have little emphasis on confusing samples. Thus, we leverage Clothing1M~\cite{xiao2015learning} fashion classification dataset for its noisy labels and inter-class similarities. The training split has a small clean and a large noisy labeled subsets. Inter-class similarity,~\eg, \textit{Down Coat} and \textit{Windbreaker}, and images with wrong labels are two distinct confusion sources, both of which are captured by our learned uncertainty model.

One of the main objectives behind modeling uncertainty is improving safety, since uncertainty quantification can prevent error propagation~\cite{mcallister2017concrete}. To this end, we employ Honda driving dataset (HDD)~\cite{RamanishkaCVPR2018} for evaluation on safety-critical autonomous navigation domain. Explicit heteroscedastic uncertainty representation improves retrieval performance by reducing the effect of noisy data with the implied attenuation. Qualitative evaluation demonstrates the ability of our approach to identify confusing driving situations.
 




In summary, the key contributions of this paper are:
\vspace{-0.1in}
\begin{enumerate}[noitemsep]
	\item Formulating an unsupervised triplet loss extension to capture heteroscedastic (data) uncertainty in visual retrieval systems.
	\item Improving retrieval model's interpretability by identifying confusing visual objects in train and test data. This reduces error propagation and enables data cleaning.
	\item Harnessing heteroscedastic uncertainty to improve efficiency by 1-2\% and improving model stability by modeling local noise in the embedding space.

\end{enumerate}

\section{Related Work}

\subsection{Bayesian Uncertainty Modeling}
Bayesian models define two types of uncertainty: epistemic and aleatoric. Epistemic uncertainty, also known as model uncertainty, captures uncertainty in model parameters. It reflects generalization error and can be reduced given enough training data. Aleatoric uncertainty is the uncertainty in our data, \eg, uncertainty due to observation noise. Kendall and Gal~\yrcite{kendall2017uncertainties} divide it into two sub-categories: heteroscedastic and homoscedastic. Homoscedastic is task-dependent uncertainty not dependent on the input space, \ie, constant for all input data and varies between different tasks. Heteroscedastic varies across the input space due to observational noise,~\ie, $\sigma(x)$.

Quantifying uncertainties can potentially improve the performance, robustness, and interpretability of a system. Therefore, epistemic uncertainty modeling has been leveraged for semantic segmentation~\cite{nair2018exploring}, depth estimation~\cite{kendall2017uncertainties}, active learning~\cite{gal2017deep}, conditional retrieval~\cite{taha2019exploring}, and model selection~\cite{gal2016dropout} though hyper-parameter tuning. A supervised approach to learning heteroscedastic uncertainty to capture observational noise has been proposed~\cite{nix1994estimating,le2005heteroscedastic}. However, labeling heteroscedastic uncertainty in real-world problems is challenging and not scalable.  

A recent approach~\cite{kendall2017uncertainties} regresses this uncertainty without supervision. This approach has been applied in semantic segmentation and depth estimation. By making the observation noise parameter $\sigma$ data-dependent, it can be learned as a function of the data $x_i$ as follows
\begin{equation}\label{eq:hetero}
L=\frac { 1 }{ N } \sum _{ i=1 }^{ N }{ \frac { 1 }{ 2{ \sigma (x_{ i }) }^{ 2 } } { \left\| y_{ i }-f(x_{ i }) \right\|  }^{ 2 }+\frac { 1 }{ 2 } \log { \sigma (x_{ i }) }  }, 
\end{equation}
for a labeled dataset with $N$ points of $(x_i,y_i)$ and $f(x_i)$ is a uni-variate regression function. 
This formulation allows the network to reduce the erroneous labels' effect. The noisy data with predicted high uncertainty will have a smaller effect on the loss function $L$ which increases the model robustness. The two terms in equation~\ref{eq:hetero} have contradicting objectives. While the first term favors high uncertainty for all points, the second term $\log(\sigma(x_i))$ penalizes it. 

We extend triplet loss to learn data uncertainty in a similar unsupervised manner. The network learns to ignore parts of the input space if uncertainty justifies penalization. This form of learned attenuation is a consequence of the probabilistic interpretation of~\cite{kendall2017uncertainties} model.

\subsection{Triplet Loss}
To learn a space embedding, we leverage triplet loss for its simplicity and efficiency. It is more efficient than contrastive loss~\cite{hadsell2006dimensionality,li2017improving}, and less computationally expensive than  quadruplet~\cite{huang2016local,chen2017beyond} and quintuplet~\cite{huang2016learning} losses.  Equation~\ref{eq:triplet} shows the triplet loss formulation 
\newcommand\sbullet[1][.5]{\mathbin{\vcenter{\hbox{\scalebox{#1}{$\bullet$}}}}}
\begin{equation}\label{eq:triplet}
L_\text{tri}=\frac{1}{b} \sum _{ i=1 }^{ b }{ { \left[  { (D(\left\lfloor a\right\rfloor,\left\lfloor p\right\rfloor)-{ D(\left\lfloor a\right\rfloor,\left\lfloor n\right\rfloor) } +m) }  \right]  }_{ + },  }
\end{equation}
where ${ \left[ \sbullet[0.75] \right]  }_{ + }$ is a soft margin function and $m$ is the margin between different classes embedding. $\left\lfloor\sbullet[0.75] \right\rfloor$ and $D(,)$ are the embedding and the Euclidean distance functions respectively. This formulation attracts an anchor image $a$ of a specific class closer to all other positive images $p$ from  the same class than it is to any negative image $n$ of other classes.

\begin{figure}[h]
	\centering
	\includegraphics[width=0.6\linewidth]{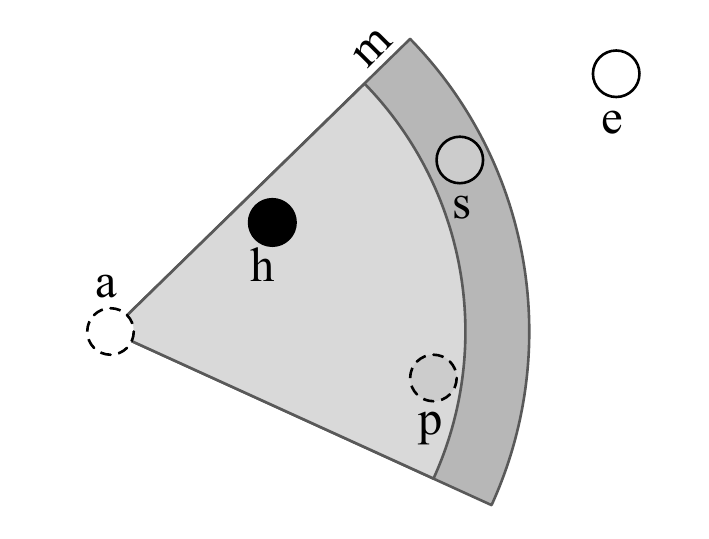}
	\caption{Triplet loss tuple (anchor, positive, negative) and margin $m$. (H)ard, (s)emi-hard and (e)asy negatives highlighted in black, gray and white respectively.}
	\label{fig:semi_neg}
\end{figure}

The performance of triplet loss relies heavily on the sampling strategy used during training. We experiment with both hard~\cite{hermans2017defense} and semi-hard sampling~\cite{schroff2015facenet} strategies. In semi-hard negative sampling, instead of picking the hardest positive-negative samples, all anchor-positive pairs and their corresponding semi-hard negatives are considered. Semi-hard negatives are further away from the anchor than the positive exemplar, yet within the banned margin $m$.  Figure~\ref{fig:semi_neg} shows a triplet loss tuple and highlights different types of negative exemplars. Hard and semi-hard negatives satisfy equations~\ref{eq:hard_neg} and~\ref{eq:semi_neg} respectively. 
\begin{align}\label{eq:semi_neg}
D(\left\lfloor a\right\rfloor,\left\lfloor p\right\rfloor) < { D( \left\lfloor a\right\rfloor,\left\lfloor n\right\rfloor) } < D( \left\lfloor a\right\rfloor,\left\lfloor p\right\rfloor ) + m, \\ \label{eq:hard_neg}
p = \arg\!\max_{i} {D(\left\lfloor a\right\rfloor,\left\lfloor i\right\rfloor)}, n = \arg\!\min_i D(\left\lfloor a\right\rfloor,\left\lfloor i\right\rfloor).
\end{align}

Triplet loss has been extended to explore epistemic (model) uncertainty~\cite{taha2019exploring}. In this paper, we propose a similar formulation to learn heteroscedastic (data) uncertainty.

\subsection{Bayesian Retrieval}
Dropout as a Bayesian approximation framework has been theoretically studied for both classification and regression problems~\cite{gal2016dropout,kendall2017uncertainties}. To extend this framework to retrieval and space embedding problems, triplet loss is cast as a regression function~\cite{taha2019exploring}. Given a training dataset containing $N$ triplets $\{(x_1,y_1,z_1),(x_2,y_2,z_2),...(x_n,y_n,z_n)\}$ and their corresponding outputs $(d_1,...,d_n)$, the triplet loss can be formulated as a trivariate regression function as follows
\begin{align}\label{eq:tri_func}
& f_\text{tri}(x_i,y_i,z_i) = d_i \in [0,2+m]\\
&={ \left[D(\left\lfloor x_i \right\rfloor ,\left\lfloor y_i \right\rfloor )-D(\left\lfloor x_i \right\rfloor,\left\lfloor z_i \right\rfloor) +m \right]  }_{ + }. \label{eq:tri_func_full}
\end{align}
 Assuming a unit-circle normalized embedding, $f_{tri}(x_i,y_i,z_i)$ outputs $d_i=0$ if $y_i,x_i \in c_i$ and $z_i \in c_j$; and $d_i=2+m$ if $z_i,x_i \in c_i$ and $y_i \in c_j$  s.t. $i \ne j$. This casting enables epistemic uncertainty learning for multi-modal conditional retrieval systems~\cite{taha2019exploring}. Inspired by this, Section~\ref{sec:approach} presents our proposed extension to capture heteroscedastic uncertainty.

\section{Heteroscedastic Embedding}\label{sec:approach}

Heteroscedastic models investigate the observation space and identify parts suffering from high noise levels. Taha~\etal~\yrcite{taha2019exploring} cast normalized ranking losses as a regression function to study epistemic uncertainty. Similarly, we extend triplet loss to learn the data-dependent heteroscedastic uncertainty. 
This helps identify noisy and confusing objects in a retrieval system, either in queries or in the search gallery.

Normalized triplet loss is cast as a trivariate regression function~\cite{taha2019exploring}. It is straight-forward to extend it for unnormalized embedding with soft margin as follows
\begin{align}\label{eq:tri_func}
	 f_\text{tri}(x_i,y_i,z_i)&= d_i \in [0,\infty)\\
   & ={ \left[D(\left\lfloor x_i \right\rfloor ,\left\lfloor y_i \right\rfloor )-D(\left\lfloor x_i \right\rfloor,\left\lfloor z_i \right\rfloor) \right]  }_{ + }. \label{eq:tri_func_full}
\end{align}
$f_\text{tri}(x_i,y_i,z_i)$ outputs $d_i=0$ if $y_i,x_i \in c_i$ and $z_i \in c_j$; and $d_i \rightarrow \infty$ if $z_i,x_i \in c_i$ and $y_i \in c_j$  s.t. $i \ne j$. Unlike the univariate regression formulation~\cite{kendall2017uncertainties}, triplet loss is dependent on three objects: anchor, positive, and negative. We extend the vanilla triplet loss to learn a noise parameter $\sigma$ for each object independently, i.e., $\sigma_a, \sigma_p, \sigma_n$. For a single triplet $(a,p,n)$, the vanilla triplet loss is evaluated three times as follows
\begin{align}\label{eq:tri_func_hetero}
\nonumber  f_\text{tri}(a,p,n)&= \frac { 1 }{ 2{ \sigma_a }^{ 2 }} L_\text{tri}(a,p,n) +\frac { 1 }{ 2 } \log { \sigma_a^2 }  \\ \nonumber
& + \frac { 1 }{ 2{ \sigma_p }^{ 2 }} L_\text{tri}(a,p,n) +\frac { 1 }{ 2 } \log { \sigma_p^2}  \\ 
& + \frac { 1 }{ 2{ \sigma_n }^{ 2 }} L_\text{tri}(a,p,n) +\frac { 1 }{ 2 } \log { \sigma_n^2} \\
\nonumber & = \frac { L_\text{tri}(a,p,n) }{ 2} \left(\frac { 1 }{ { \sigma_a }^{ 2 }} + \frac { 1 }{ { \sigma_p }^{ 2 }}+ \frac { 1 }{ { \sigma_n }^{ 2 }}\right) \\
& +	 \frac {1}{2}\log {\sigma_a^2 \sigma_p^2 \sigma_n^2},
\end{align}

where $L_\text{tri}(a,p,n) = { \left[D(\left\lfloor a \right\rfloor ,\left\lfloor p \right\rfloor )-D(\left\lfloor a \right\rfloor,\left\lfloor n \right\rfloor) \right]  }_{ + }$.  This formulation can be regarded as a weighted average triplet loss using data uncertainty. Similar to~\cite{kendall2017uncertainties}, we compute a maximum a posteriori probability (MAP) estimate by adding a weight decay term parameterized by $\lambda$. This imposes a prior on the model parameters and reduces overfitting~\cite{le2005heteroscedastic}. Our neural network learns $s=\log{\sigma^2}$ because it is more numerically stable than regressing the variance $\sigma^2$. Thus, in practice the final loss function is 
\begin{align}
\nonumber L =   & \frac{1}{N} \sum_{(a,p,n)\in\mathcal{T}}
{\left[\frac{(e^{-s_a}+e^{-s_p}+e^{-s_n})L_{Tri}(a,p,n)}{2} \right. } \\
& \left. + \frac{(s_a+s_p+s_n)}{2}\right] + \lambda {\left\| W \right\|}^2,
\end{align}
where $N$ is the number of triplets $(a,p,n)\in\mathcal{T}$. Our formulation can be generalized to support more complex ranking losses like quintuplet loss~\cite{huang2016learning}. Equation~\ref{eq:k_tup_func} provides a generalization for k-tuplets where $k \ge 3$
\begin{align}\label{eq:k_tup_func}
& f_{k\_\text{tup}}(x_0,..,x_j,..,x_k)= \frac{L_\text{tri}(a,p,n)}{2} \left(\sum_{j=0}^{k}{\frac { 1 }{ { \sigma_j }^{ 2 }}}\right) \nonumber \\
&\qquad\qquad\qquad\qquad\qquad +	 \frac {1}{2}\log {\prod_{j=0}^{k}\frac { 1 }{ { \sigma_j }^{ 2 }}}\\
& \text{s.t.} \quad D\left(\left\lfloor x_{ 0 } \right\rfloor ,\left\lfloor x_{ 1 } \right\rfloor \right) < D(\left\lfloor x_{ 0 } \right\rfloor ,\left\lfloor x_{ j } \right\rfloor ) <  D(\left\lfloor x_{ 0 } \right\rfloor ,\left\lfloor x_{ k } \right\rfloor ).
\end{align}

\section{Architecture}
\begin{figure}[t!]
	\begin{center}
		\includegraphics[width=0.7\linewidth]{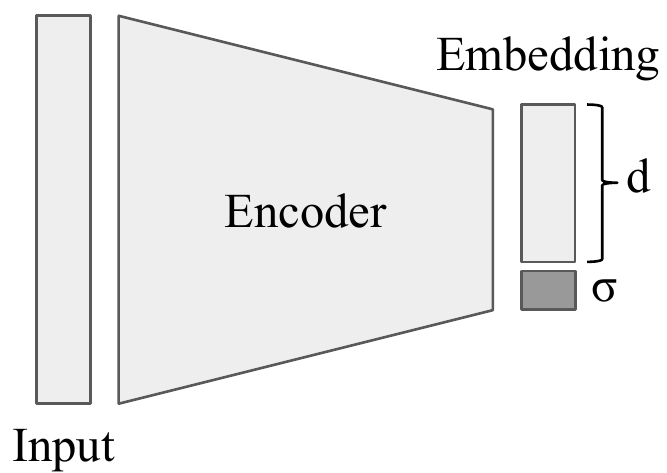}
	\end{center}
	\caption{Our generic retrieval network supports various encoder architectures trained through a ranking loss.}
	\vspace{-0.1in}
	\label{fig:arch}
\end{figure}

The generic architecture employed in our experiments is illustrated in Figure~\ref{fig:arch}. The encoder architecture is dependent on the input type. For an embedding space with dimensionality $d$, our formulation requires the encoder final layer output $\in R^{d+1}$. The extra dimension learns the input heteroscedastic uncertainty $\sigma(x)$. The following subsections present two encoder variants employed to properly handle image and video inputs.



\subsection{Image Retrieval}
For image-based tasks of person re-identification and fashion item retrieval, we employ the architecture from~\cite{hermans2017defense}. Given an input RGB image, the encoder is a fine-tuned ResNet architecture~\cite{he2016deep} pretrained on ImageNet~\cite{deng2009imagenet} followed by a fully-connected network (FCN). In our experiments, the final output is not normalized and the soft margin between classes is imposed by the softplus function $\ln(1+\exp( \sbullet[.75] ))$. It is similar to the hinge function $\max(\sbullet[.75] ,0)$ but it decays exponentially instead of a hard cut-off. We experiment with both hard and semi-hard negative sampling strategies for person re-identification and fashion item retrieval respectively.
\subsection{Video Retrieval}

For autonomous navigation, a simplified version of~\cite{taha2019exploring} architecture is employed. The Honda driving dataset provides multiple input modalities,~\eg, camera and CAN sensors, and similarity notions between actions (events). We employ the camera modality and two similarity notions: \textit{goal-oriented} and \textit{stimulus-driven}. Input video events from the camera modality are represented using pre-extracted features per frame, from the $\texttt{Conv2d\_7b\_1x1}$ layer of InceptionResnet-V2~\cite{szegedy2017inception} pretrained on ImageNet, to reduce GPU memory requirements.


Modeling temporal context provides an additional and important clue for action understanding~\cite{simonyan2014two}. Thus, the encoder employs an LSTM~\cite{funahashi1993approximation,hochreiter1997long} after a shallow CNN. During training, three random consecutive frames are drawn from an event. They are independently encoded then temporally fused using the LSTM. Note that sampling more frames per event will lead to better performance. Unfortunately, the GPU memory constrains the number of sampled frames to three. The network output is the hidden state of the LSTM last time step. Further architectural details are described in the supplementary material.



\section{Experiments}
We evaluate our formulation on three retrieval domains through three datasets. First, it is validated using the standard person-identification benchmark~\cite{zheng2017unlabeled}. In order to model inter-class similarity and local noise, we leverage two real-world datasets: Clothing1M and Honda Driving Dataset (HDD). Both datasets emulate real scenarios with noisy labels due to inter-class similarity or inherent uncertainty.

\subsection{Person Re-Identification}

Person re-identification is employed in Multi-Target Multi-Camera Tracking system. An ID system retrieves images of people and ranks them by decreasing similarity to a given person query image. DukeMTMC-reID~\cite{zheng2017unlabeled} is used for evaluation. It includes 1,404 identities appearing in more than two cameras and 408 identities appearing in a single camera for distraction purpose. 702 identities are reserved for training and 702 for testing. We evaluate our formulation using this clean dataset for two reasons: (1) It provides an established quantitative benchmark to emphasize the competence and robustness of our approach; (2) qualitative results comprehension requires no domain-specific knowledge.



For each training mini-batch, we uniformly sample $P = 18$ person identities without replacement.  For each person, $K = 4$ sample images are drawn without replacement with resolution $256\times 128$. The learning rate is $3*10^{-4}$ for the first 15000 iterations and decays to $10^{-7}$ at iteration 25000. Weight regularization employed with $\lambda=10e-4$. Our formulation is evaluated twice with and without data augmentation. Similar to~\cite{ristani2018features}, we augment images by cropping and horizontal flipping. For illumination invariance, contrast normalization, grayscale and color multiplication effects are applied. For resolution invariance, we apply Gaussian blur of varying $\sigma$. For additional viewpoint/pose invariance, we apply perspective transformations and small distortions. We additionally hide small rectangular image patches to simulate occlusion.

Figure~\ref{fig:duke_confusin_person} shows the five lowest and highest uncertainty query-identities from the DukeMTMC-ReID dataset. Heteroscedastic uncertainty is high when the query image contains multiple identities or a single identity with an outfit that blends with the background. On the contrary, identities with discriminative outfit colors (\eg, red) suffer low uncertainty. 


Table~\ref{tbl:duke} presents our quantitative evaluation where the performance of our method is comparable to the state-of-the-art. All experiments are executed five times, and mean average precision (mAP) and standard deviation are reported. Our formulation lags marginally due to limited confusing samples in the training split. However, it has a smaller standard deviation. It is noteworthy that the performance gap between vanilla Tri-ResNet and our formulation closes when applying augmentation. This aligns with our hypothesis that the lack of confusing samples limits our formulation. In the next subsections, we evaluate on real-world datasets containing noisy samples.

\begin{table}[t]
	\scriptsize
	\centering
	\caption{Quantitative evaluation on DukeMTMC-ReID}
	\begin{tabular}{@{}l c c@{}}
\toprule
		Method          & mAP & Top-5   \\ 
		\midrule
		BoW+KISSME~\cite{zheng2015scalable}    &   12.17 & -	 \\ 		LOMO+XQDA~\cite{liao2015person}    &   17.04 	& -  \\
		Baseline~\cite{zheng2016person}    &   44.99 	& -  \\
		PAN~\cite{zheng2018pedestrian}    &   51.51 	& - \\
		SVDNet~\cite{sun2017svdnet}    &   56.80 	 & - \\
		\midrule
		Tri-ResNet~\cite{hermans2017defense}    &  \textbf{ 56.08$\pm$0.005} & \textbf{86.76$\pm$0.007}  \\ 
			\textbf{Tri-ResNet + Hetero (ours)}   &    55.16$\pm$0.002	&  86.03$\pm$0.005\\
					\midrule

Tri-ResNet + Aug    &   56.44$\pm$0.006 &  86.11$\pm$0.003 \\ 
\textbf{Tri-ResNet + Aug + Hetero (ours)}   &   \textbf{ 56.74$\pm$0.004} & \textbf{86.20$\pm$0.002} \\ 
\bottomrule

	\end{tabular}
	\vspace{-0.2in}
	\label{tbl:duke}
\end{table}

\subsection{Fashion Image Retrieval}
A major drawback of the person re-identification dataset is the absence of noisy images. Images with multiple identities are confusing but incidental in the training split. To underscore the importance of our formulation, a large dataset with noisy data is required. Clothing1M fashion dataset~\cite{xiao2015learning} emulates this scenario by providing a large-scale dataset of clothing items crawled from several online shopping websites. It contains over one million images and their description. Fashion items are labeled using a noisy process: a label is assigned if the description contains the keywords of that label, otherwise, it is discarded. A small clean portion of the data is made available after manual refinement. The final training split contains 22,933 clean ($D_c$) and 1,024,637 (97.81\%) noisy ($D_n$) labeled images. The validation and test sets have $14,313$ and $10,526$ clean images respectively.

\begin{figure}[t]
	\begin{tikzpicture} \begin{axis}[ ybar,width=0.5\textwidth, height=3.0cm, enlargelimits=0.15, symbolic x coords={T-Shirt, Shirt, Knitwear, Chiffon, Sweater, Hoodie, Windbreaker, Jacket, Down Coat, Suit, Shawl, Dress, Vest, Underwear}, xtick=data, nodes near coords align={vertical}, x tick label style={rotate=45,anchor=east},] 
	\addplot coordinates {
		(T-Shirt,92879)
		(Shirt,93077) 
		(Knitwear,90361)
		(Chiffon,56872)
		(Sweater,24442)
		(Hoodie,88135) 
		(Windbreaker,84886)
		(Jacket,83365) 
		(Down Coat,79544) 
		(Suit,93415)
		(Shawl,45709)
		(Dress,93666)
		(Vest,65550)
		(Underwear,80508)
	};  \end{axis}
	\end{tikzpicture}
		\vspace{-0.1in}
	\caption{Clothing1M classes distribution.}
		\vspace{-0.05in}
	\label{fig:clothing_dist}
\end{figure}
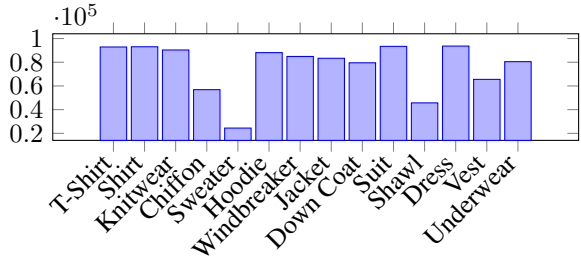

Figure~\ref{fig:clothing_dist} shows the 14 classes, and their distribution, from the Clothing1M dataset. Training with both clean and noisy is significantly superior compared to just training with clean data. Since manual inspection of noisy data is expensive, our unsupervised formulation qualitatively identifies confusing samples and provides an efficient system to deal with noisy data. For clothing1M dataset, the training parameters are similar to person re-identification dataset except for the following: a minibatch has samples from all 14 classes ($k=10$ samples per class per minibatch) and input image resolution is $224\times 224$. Semi-hard sampling strategy is employed to mitigate noisy labels effect. The model is training for 25k training iterations, which is equivalent to six epochs.

\begin{table}[t]
	\scriptsize
	\centering
	\caption{Quantitative evaluation on Clothing1M. The first row show performance with clean data only while the remaining rows leverage both clean and noisy data. The last two rows show performance after cleaning 20\% of the search gallery samples}
	\begin{tabular}{@{}l c@{}}
		\toprule
		Method          & mAP  \\
		\midrule
		Tri-ResNet (Clean Only)    &   52.62  \\
				\midrule 
		Tri-ResNet  (Baseline)  &   61.70$\pm$0.001  \\ 
		\textbf{Tri-ResNet + Hetero (ours)}   &   \textbf{ 62.27$\pm$0.001} \\ 		\midrule
		
		Tri-ResNet  + Random Cleaning  &   62.33$\pm$0.003  \\
		\textbf{Tri-ResNet + Hetero Cleaning}   &   \textbf{ 64.57$\pm$0.002} \\ \bottomrule 
		
	\end{tabular}
	\label{tbl:cloth_quan}
		\vspace{-0.1in}
\end{table}
%
%

The validation and test splits act as query and gallery database respectively for quantitative evaluation. Table~\ref{tbl:cloth_quan} presents retrieval performance using only clean data \emph{vs.}\ both clean and noisy data. Mean and standard deviation across five trails are reported. By modeling data uncertainty, our formulation improves performance and interpretability of the model. We utilize the learned uncertainty to clean confusing samples from the search gallery database. The last two rows in Table~\ref{tbl:cloth_quan} present retrieval performance after cleaning 20\% of the search database. While random cleaning achieves no improvement, removing items suffering the highest uncertainty boosts performance by $\sim$2\%.

\newcommand*\rot{\rotatebox{90}}
\begin{figure}[t!]
	\centering
	\setlength\tabcolsep{0.5pt} 
	\setlength{\fboxsep}{0pt}%
	\setlength{\fboxrule}{2pt}%
	\begin{tabular}{@{}cccccc@{}}
		\addvbuffer[1.0ex]{\rot{Sweater}}&
		\fcolorbox{white}{white}{\includegraphics[width=0.08\textwidth,height=0.08\textwidth]{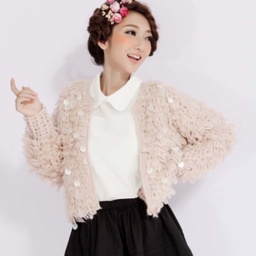}} &
		\fcolorbox{white}{white}{\includegraphics[width=0.08\textwidth,height=0.08\textwidth]{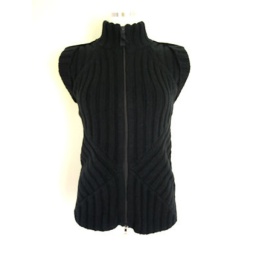}} & \fcolorbox{white}{white}{\includegraphics[width=0.08\textwidth,height=0.08\textwidth]{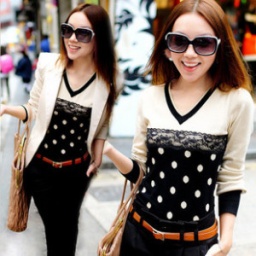}} & 
		\fcolorbox{white}{white}{\includegraphics[width=0.08\textwidth,height=0.08\textwidth]{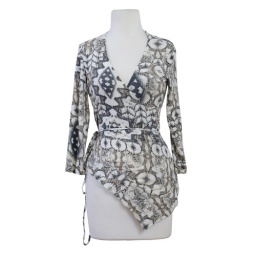}} &
		\fcolorbox{white}{white}{\includegraphics[width=0.08\textwidth,height=0.08\textwidth]{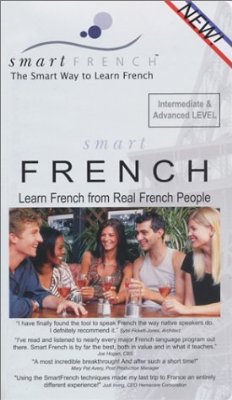}}  \\ \hline
		\addvbuffer[1.5ex]{\rot{Suit}}&
		\fcolorbox{white}{white}{\includegraphics[width=0.08\textwidth,height=0.08\textwidth]{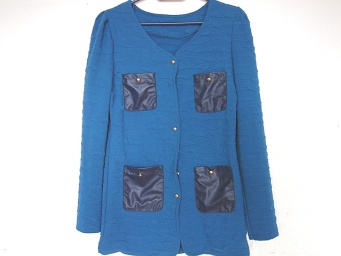}} &
		\fcolorbox{white}{white}{\includegraphics[width=0.08\textwidth,height=0.08\textwidth]{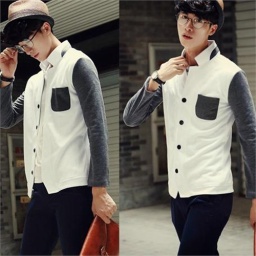}} & \fcolorbox{white}{white}{\includegraphics[width=0.08\textwidth,height=0.08\textwidth]{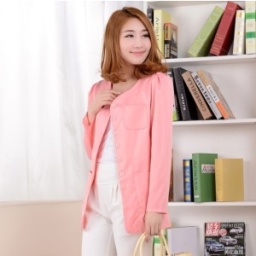}} & 
		\fcolorbox{white}{white}{\includegraphics[width=0.08\textwidth,height=0.08\textwidth]{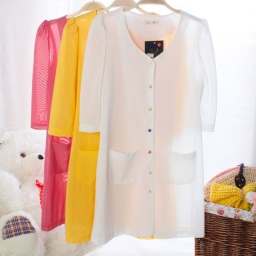}} &
		\fcolorbox{white}{white}{\includegraphics[width=0.08\textwidth,height=0.08\textwidth]{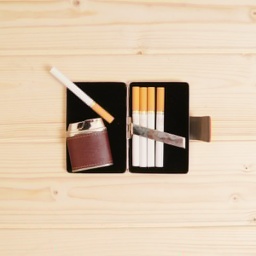}} \\
	\end{tabular}
	\caption{Qualitative evaluation using the highest five uncertainty training images from \textit{Sweater} and \textit{Suit} classes in Clothing1M.}
	\label{fig:clothing_qual_trn_cleanse_main}
	\vspace{-0.1in}
\end{figure}

Leveraging uncertainty to refine the training split is a plausible extension but requires extensive manual labor. Figure~\ref{fig:clothing_qual_trn_cleanse_main} presents the five highest uncertainty training images from two classes. The supplementary material provides a qualitative evaluation showing images with highest uncertainty score from each class. Most images are either incorrectly labeled or contain multiple distinct objects which highlights the utility of our approach.

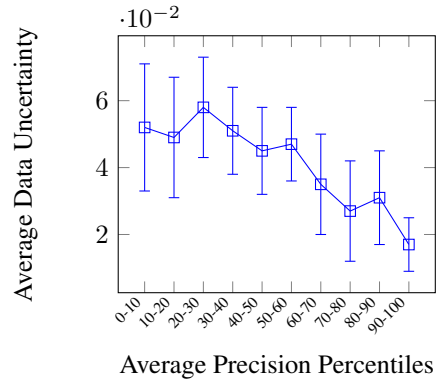
\begin{figure}[h!]
	\centering
	\begin{tikzpicture}
	\begin{axis}[
	height=5.0cm,
	symbolic x coords={0-10,10-20,20-30,30-40,40-50,50-60,60-70,70-80,80-90,90-100},
	x tick label style={rotate=45,anchor=east,font=\tiny},
	xtick=data,
	xlabel={Average Precision Percentiles},
	x label style={at={(axis description cs:0.5,-0.05)},anchor=north},
	ylabel={Average Data Uncertainty},
	]
	
	\addplot[
	color=blue,
	mark=square,
	error bars/.cd,
	y dir=both,y explicit
	]
	coordinates {
		(0-10,0.052) +- (0,0.019) 
		(10-20,0.049) +- (0,0.018) 
		(20-30,0.058) +- (0,0.015)
		(30-40,0.051) +- (0,0.013)
		(40-50,0.045) +- (0,0.013)
		(50-60,0.047) +- (0,0.011)
		(60-70,0.035) +- (0,0.015)
		(70-80,0.027) +- (0,0.015)
		(80-90,0.031) +- (0,0.014)
		(90-100,0.017) +- (0,0.008)
	};
	
	\end{axis}
	\end{tikzpicture}
		\vspace{-0.05in}
	\caption{Quantitative analysis reveals the $-ve$ correlation between query images retrieval average precision and uncertainty. Queries with high average precision suffer lower uncertainty and vice versa. Query images are aggregated using average precision percentiles on the x-axis. Y-axis is the aggregated images' mean uncertainty and standard deviation.}
	\label{fig:clothing1m_correlation}
\end{figure}

\begin{figure}[h!]
	\setlength\tabcolsep{0.5pt} 
	\setlength{\fboxsep}{0pt}%
	\setlength{\fboxrule}{2pt}%
	\begin{tabular}{@{}c|cccc@{}}
		
		Query & \multicolumn{4}{c}{Top 4 results}
		\\ \hline
		\fcolorbox{red}{white}{\includegraphics[width=0.08\textwidth,height=0.08\textwidth]{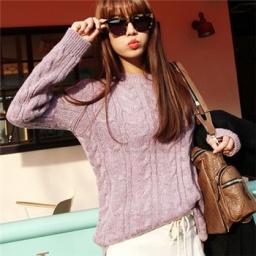}} &
		\fcolorbox{red}{white}{\includegraphics[width=0.08\textwidth,height=0.08\textwidth]{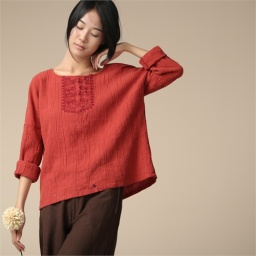}} & \fcolorbox{violet}{white}{\includegraphics[width=0.08\textwidth,height=0.08\textwidth]{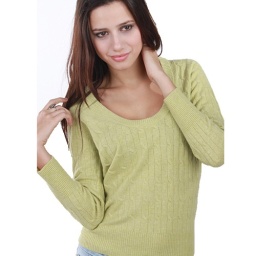}} & 
		\fcolorbox{red}{white}{\includegraphics[width=0.08\textwidth,height=0.08\textwidth]{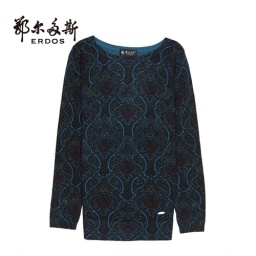}} &
		\fcolorbox{red}{white}{\includegraphics[width=0.08\textwidth,height=0.08\textwidth]{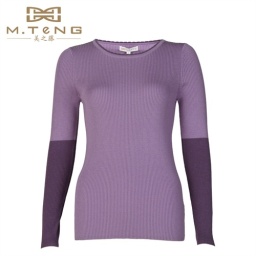}} 
		\\ 
		Sweater & Knitwear & Sweater & Sweater & Sweater
		\\ \hline
		\fcolorbox{red}{white}{\includegraphics[width=0.08\textwidth,height=0.08\textwidth]{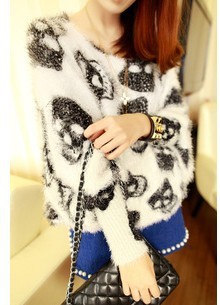}} &
		\fcolorbox{red}{white}{\includegraphics[width=0.08\textwidth,height=0.08\textwidth]{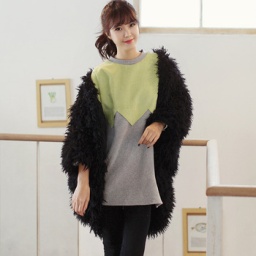}} & \fcolorbox{red}{white}{\includegraphics[width=0.08\textwidth,height=0.08\textwidth]{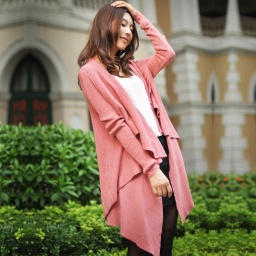}} & 
		\fcolorbox{red}{white}{\includegraphics[width=0.08\textwidth,height=0.08\textwidth]{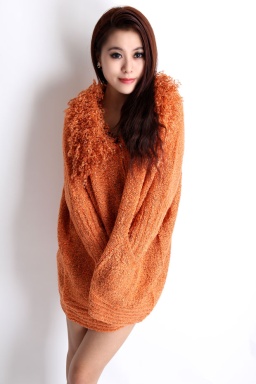}} &
		\fcolorbox{violet}{white}{\includegraphics[width=0.08\textwidth,height=0.08\textwidth]{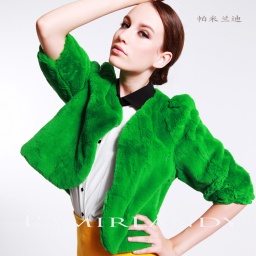}}
		\\ 
		Knitwear & Windbreaker & Windbreaker & Windbreaker & Shawl
		\\ \hline
		\fcolorbox{red}{white}{\includegraphics[width=0.08\textwidth,height=0.08\textwidth]{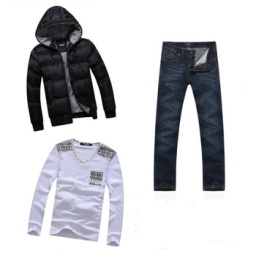}} &
		\fcolorbox{violet}{white}{\includegraphics[width=0.08\textwidth,height=0.08\textwidth]{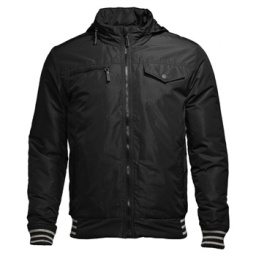}} & \fcolorbox{violet}{white}{\includegraphics[width=0.08\textwidth,height=0.08\textwidth]{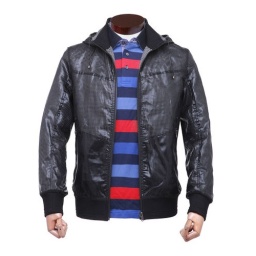}} & 
		\fcolorbox{violet}{white}{\includegraphics[width=0.08\textwidth,height=0.08\textwidth]{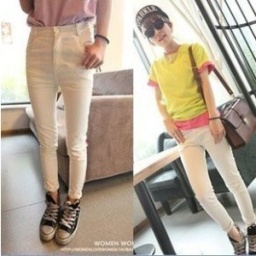}} &
		\fcolorbox{yellow}{white}{\includegraphics[width=0.08\textwidth,height=0.08\textwidth]{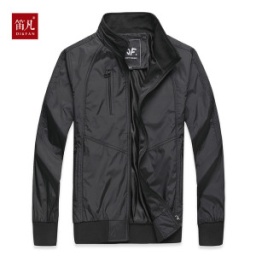}} 				\\
		Jacket & Down Coat & Down Coat & T-Shirt & Jacket
	\end{tabular}
	\caption{Qualitative evaluation using three very high uncertainty queries from Clothing1M dataset. Outline colors emphasize the uncertainty degree,~\eg, red is very high. Inter-class similarity is a primary confusion source.}
	\label{fig:qual_clothing1m_high}
	\vspace{-0.05in}
\end{figure}

Figure~\ref{fig:clothing1m_correlation} depicts a negative Pearson correlation $(r=-0.5001)$ between the retrieval average precision of query items and their heteroscedastic uncertainty. Query images are aggregated by the average-precision percentiles on the x-axis. Aggregated items' average uncertainty is reported on the y-axis. Figure~\ref{fig:qual_clothing1m_high} shows query images chosen from the $1^\text{st}$ highest uncertainty percentile and their corresponding four top results. For visualization purposes, we discretize the data uncertainty using percentiles into five bins: very low (green), low (yellow), moderate (orange), high (violet), and very high (red). Confusion between certain classes like \textit{Sweater} and \textit{Knitwear}, \textit{Knitwear} and \textit{Windbreaker}, and \textit{Jacket} and \textit{Down coat} is evident.

\begin{figure*}[ht!]
	\includegraphics[width=1.0\linewidth]{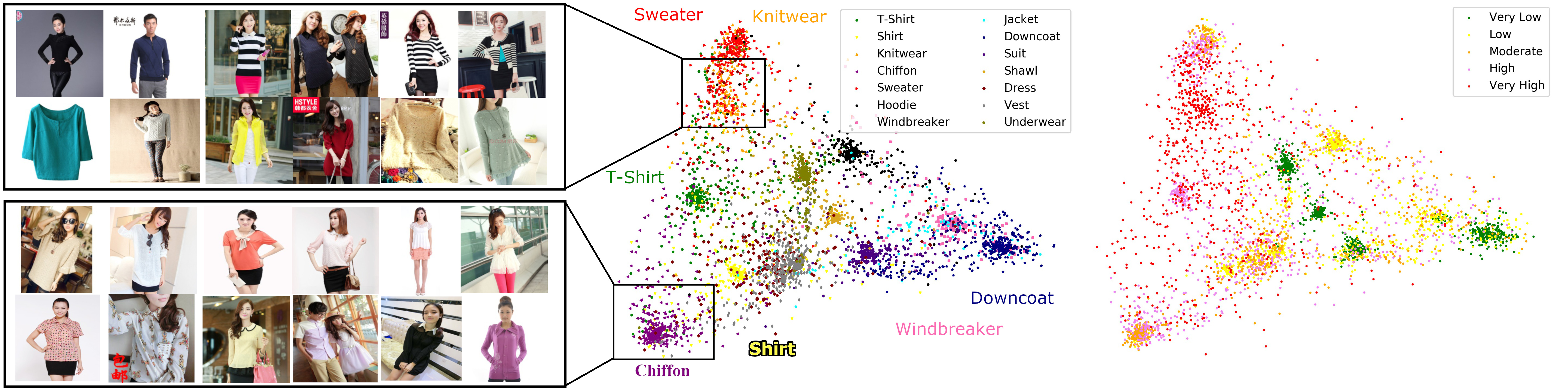}
\caption{Qualitative analysis for the Clothing1M dataset embedding using 4K random points. The left and right plots show a $2D$ PCA projection colored with class-label and uncertainty degree respectively. Points closer to class centers suffer lower uncertainty compared to farther points. Confusing inter-class similarity is highlighted with visual samples. The left zoom-in figures show four rows with samples from \textit{Sweater}, \textit{Knitwear}, \textit{Chiffon}, and \textit{Shirt} respectively. These high-resolution figures are best viewed in color/screen.}
\vspace{-0.05in}
\label{fig:clothing_uncertainty_corelation}
\end{figure*}

Figure~\ref{fig:clothing_uncertainty_corelation} shows a principal component analysis (PCA) $2D$ projection for 4K randomly chosen query items embedding. Points in the left and right projections are colored by the class label and uncertainty degree respectively. Images at the center of classes (in green) have lower uncertainty, compared to points spread out through the space. The inherent inter-class similarity, \eg, \textit{Sweater} and \textit{Knitwear}, explains why certain regions have very higher uncertainty. Qualitative evaluation with very low uncertainty query items is provided in the supplementary material.

\subsection{Autonomous Navigation}

\begin{figure}[h!]
	\begin{tikzpicture} \begin{axis}[ ybar,width=0.5\textwidth, height=3.0cm, enlargelimits=0.15, 
	symbolic x coords={
		Stop 4 sign,
		Stop 4 light,
		Stop 4 congestion,
		Stop for others,
		Stop 4 pedestrian,
		Avoid parked car
	}, xtick=data, nodes near coords align={vertical}, x tick label style={rotate=30,anchor=east},] 
	\addplot coordinates {
		(Stop 4 sign,2322)
		(Stop 4 light,754) 
		(Stop 4 congestion,1943)
		(Stop for others,50)
		(Stop 4 pedestrian,100)
		(Avoid parked car,140) 
	};  \end{axis}
	\end{tikzpicture}
		\vspace{-0.15in}
	\caption{HDD imbalance stimulus-driven actions distribution.}
	\label{fig:stimuli_dist}
	\vspace{-0.05in}
\end{figure}
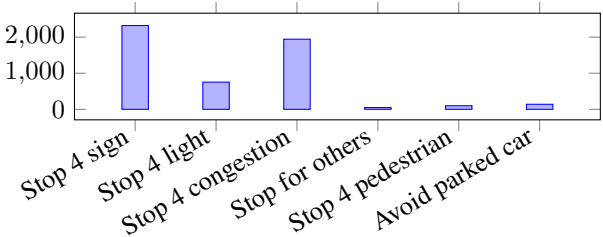

Modeling network and data uncertainty is gaining momentum in safety-critical domains like autonomous driving. We evaluate our approach on ego-motion action retrieval. Honda driving dataset (HDD)~\cite{RamanishkaCVPR2018} is designed to support modeling driver behavior and understanding causal reasoning. It defines four annotation layers. \textbf{(1) Goal-oriented actions} represent the egocentric activities taken to reach a destination like left and right turns. \textbf{(2) Stimulus-driven} are actions due to external causation factors like stopping to avoid a pedestrian or stopping for a traffic light. \textbf{(3) Cause} indicates the reason for an action. Finally, the \textbf{(4) attention} layer localizes the traffic participants that drivers attend to. Every layer is categorized into a set of classes (actions). Figures~\ref{fig:HDD_dist} and~\ref{fig:stimuli_dist} show the class distribution for goal-oriented and stimulus-driven layers respectively.

\begin{figure}[h!]
	\begin{tikzpicture} \begin{axis}[ ybar,width=0.5\textwidth, height=3.0cm, enlargelimits=0.15, symbolic x coords={Intersection Pass,Left Turn,Right Turn,Left Lane Change,Right Lane Change,Cross-walk passing,U-Turn,Left Lane Branch,Right Lane Branch,Merge}, xtick=data, nodes near coords align={vertical}, x tick label style={rotate=30,anchor=east},] 
	\addplot coordinates {
		(Intersection Pass,5651)
		(Left Turn,1689) 
		(Right Turn,1677)
		(Left Lane Change,560)
		(Right Lane Change,518)
		(Cross-walk passing,182) 
		(U-Turn,85)
		(Left Lane Branch,235) 
		(Right Lane Branch,93) 
		(Merge,143)
	};  \end{axis}
	\end{tikzpicture}
		\vspace{-0.25in}
	\caption{HDD long tail goal-oriented actions distribution.}
	\vspace{-0.05in}
	\label{fig:HDD_dist}
\end{figure}
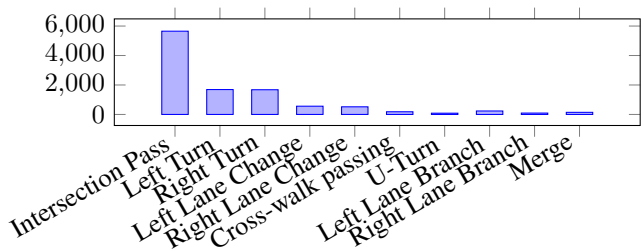

Experiments' technical details are presented in the supplementary material. An event retrieval evaluation using query-by-example is performed. Given a query event, similarity scores to all events are computed,~\ie, a leave-one-out cross evaluation on the test split. Performances of all queries are averaged to obtain the final evaluation.

To tackle data imbalance and highlight performance on minority classes, both micro and macro average accuracies are reported. Macro-average computes the metric for each class independently before taking the average. Micro-average is the traditional mean for all samples. Macro-average treats all classes equally while micro-averaging favors majority classes. Tables~\ref{tbl:goal_lbl} and~\ref{tbl:stimuli_lbl} show quantitative evaluation for networks trained on goal-oriented and stimulus-driven events respectively.

\begin{table}[h!]
	\scriptsize
	\centering
		\caption{Quantitative evaluation on goal-oriented actions}
	\begin{tabular}{@{}l c c}
		\toprule
		Method            & Baseline & Hetero (Our)\\ \midrule
		Micro mAP               &     77.88$\pm$0.003     &  \textbf{78.45$\pm$0.004}        \\
		Macro mAP         &     \textbf{32.62$\pm$0.004}     &   30.8$\pm$0.004     \\ \midrule
		Intersection Passing       &     89.1$\pm$0.006   &   \cellcolor{gray!40}91.44$\pm$0.002   \\
		Left turn      & 		    \cellcolor{gray!40}81.29$\pm$0.007   & 		  80.15$\pm$0.008   \\
		Right Turn &     \cellcolor{gray!40}89.99$\pm$0.008     &   89.19$\pm$0.016     \\
		Left Lane Change     &     \cellcolor{gray!40}24.65$\pm$0.005    &   20.28$\pm$0.008   \\
		Right Lane Change &   \cellcolor{gray!40}16.04$\pm$0.018   &  9.19$\pm$0.002   \\
		Crosswalk Passing  &       \cellcolor{gray!40}1.13$\pm$0.001  &  1.31$\pm$0.002     \\
		U-turn  &       \cellcolor{gray!40}3.59$\pm$0.004  &  2.62$\pm$0.003     \\
		Left Lane Branch  &       \cellcolor{gray!40}14.03$\pm$0.022   &  8.65$\pm$0.024     \\
		Right Lane Branch  &       \cellcolor{gray!40}2.15$\pm$0.001   &  1.48$\pm$0.011     \\
		Merge  &       \cellcolor{gray!40}4.26$\pm$0.005   &  3.62$\pm$0.006     \\ \bottomrule
	\end{tabular}
	\label{tbl:goal_lbl}
\end{table}

\begin{table}[h!]
	\scriptsize
	\centering
		\caption{Quantitative evaluation on stimulus-driven actions}
	\begin{tabular}{@{}l c c}
		\toprule
		Method            & Baseline & Hetero (Our)\\
		\midrule
		Micro mAP               &     66.50$\pm$0.008     &  \textbf{68.20$\pm$0.008}        \\
		Macro mAP         &     35.33$\pm$0.005     &   \textbf{36.23$\pm$0.008}     \\ \midrule 
		Stop  4 Sign       &     87.85$\pm$0.005   &   \cellcolor{gray!40}89.18$\pm$0.006   \\ 
		Stop 4 Light      & 		   \cellcolor{gray!40} 52.63$\pm$0.013   & 		  49.86$\pm$0.005   \\
		
		Stop 4 Congestion &     63.88$\pm$0.014     & \cellcolor{gray!40}  67.88$\pm$0.016     \\
		
		Stop 4 Others     &    \cellcolor{gray!40} 1.62$\pm$0.010    &    1.02$\pm$0.004    \\
		
		Stop 4 Pedestrian &  \cellcolor{gray!40} 2.72$\pm$0.007    &  2.56$\pm$0.002   \\
		
		Avoid Parked Car  &       3.30$\pm$0.002   &  \cellcolor{gray!40} 6.90$\pm$0.024     \\ \bottomrule
		
	\end{tabular}
	\label{tbl:stimuli_lbl}
\end{table}

%
%
%
%
%

\begin{figure*}[ht!]
	\centering
	\begin{subfigure}{1.0\textwidth}
		\centering
		\setlength{\fboxsep}{0pt}%
		\setlength{\fboxrule}{2pt}%
		\fcolorbox{red}{white}{\includegraphics[width=.19\linewidth]{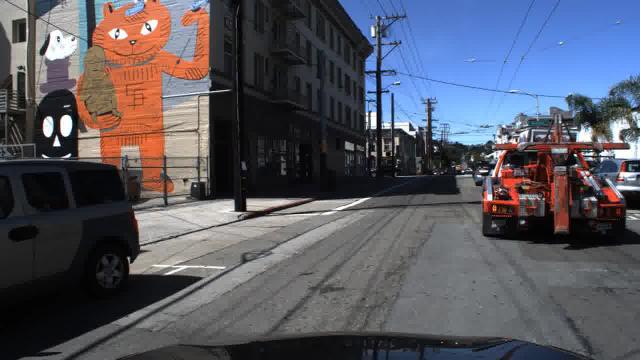}
			\includegraphics[width=.19\linewidth]{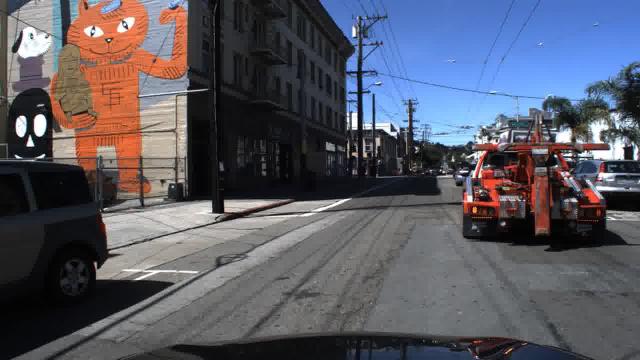}
			\includegraphics[width=.19\linewidth]{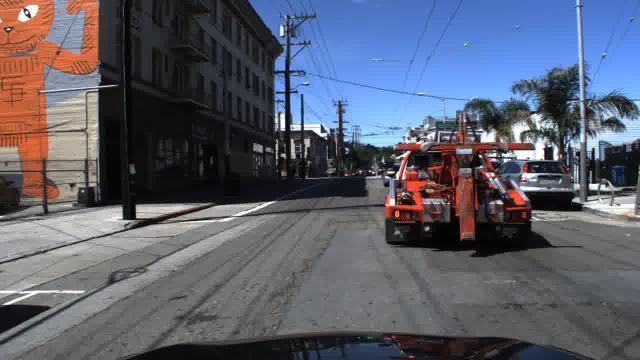}
			\includegraphics[width=.19\linewidth]{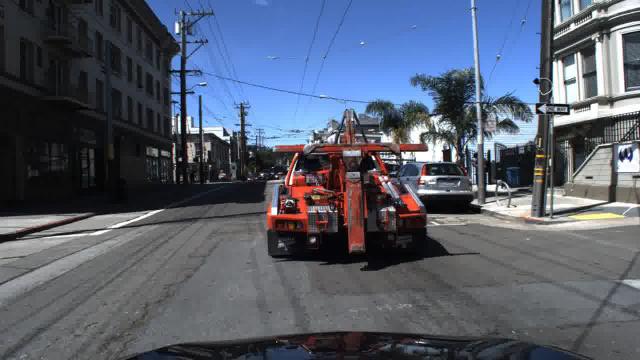}
			\includegraphics[width=.19\linewidth]{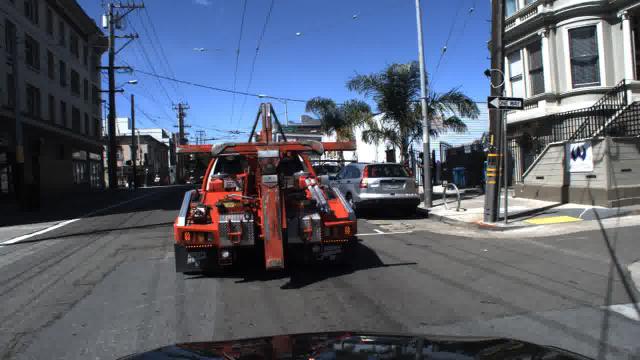}}\\	
			(Query) Right Lane Change with very high uncertainty 
	\end{subfigure}
	\begin{subfigure}{1.0\textwidth}
		\centering
		\setlength{\fboxsep}{0pt}%
		\setlength{\fboxrule}{2pt}%
		\fcolorbox{red}{white}{\includegraphics[width=.19\linewidth]{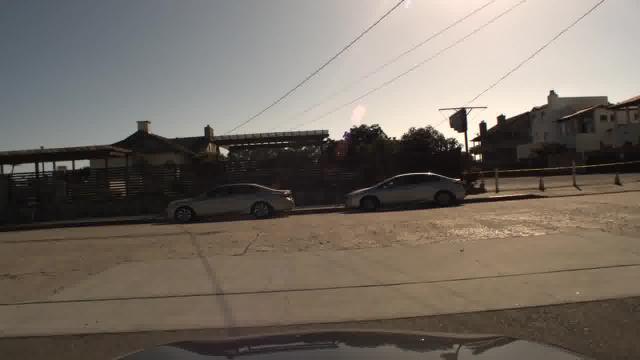}
			\includegraphics[width=.19\linewidth]{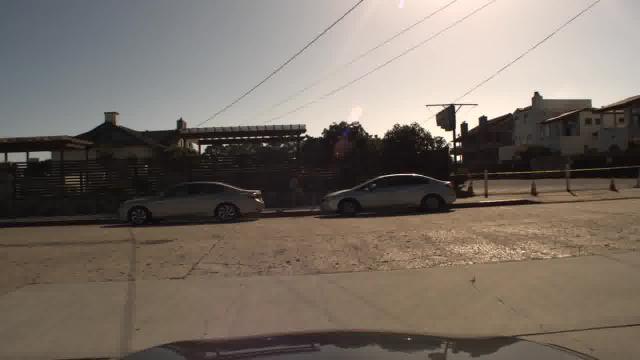}
			\includegraphics[width=.19\linewidth]{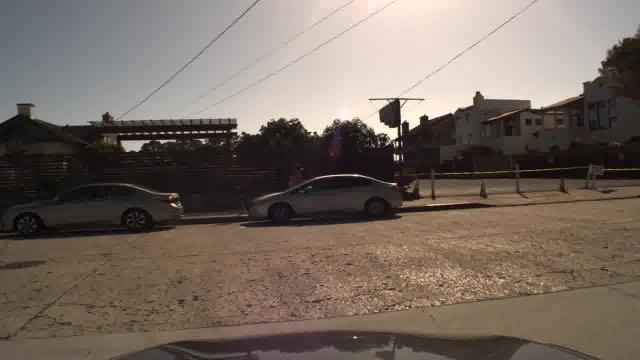}
			\includegraphics[width=.19\linewidth]{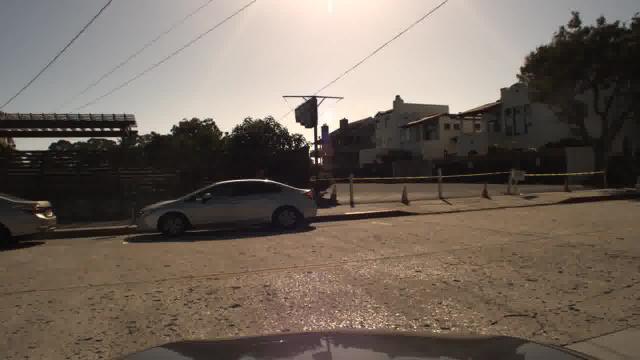}
			\includegraphics[width=.19\linewidth]{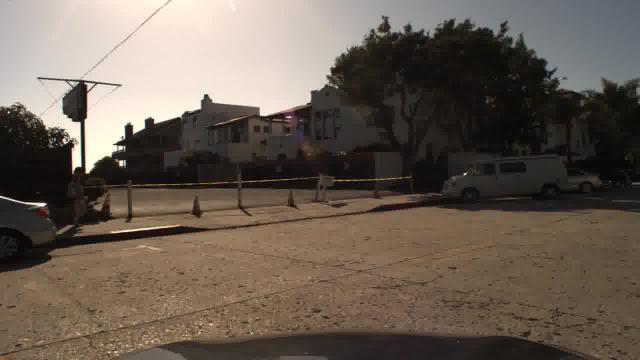}} \\ 
			(Retrieval result) Right Turn with very high uncertainty
	\end{subfigure}
	\rule{1.0\textwidth}{2.0pt}
	\begin{subfigure}{1.0\textwidth}
		\centering
		\setlength{\fboxsep}{0pt}%
		\setlength{\fboxrule}{2pt}%
		\fcolorbox{red}{white}{\includegraphics[width=.19\linewidth]{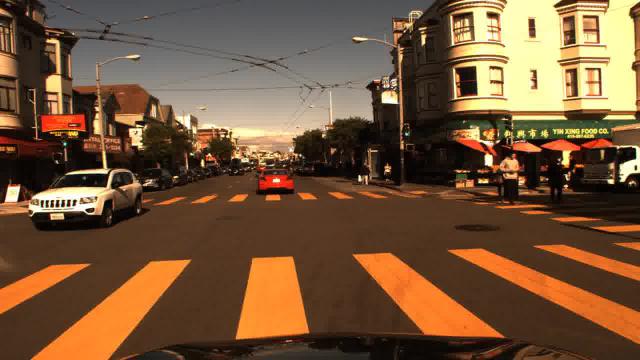}
			\includegraphics[width=.19\linewidth]{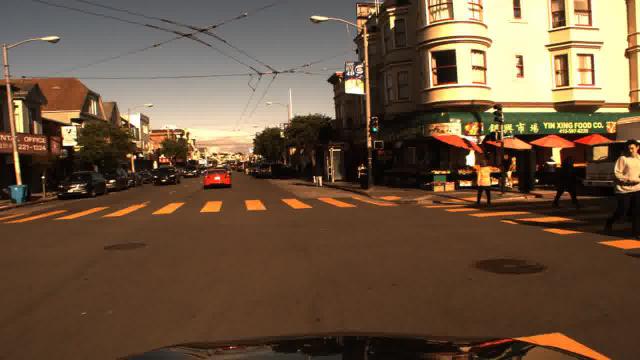}
			\includegraphics[width=.19\linewidth]{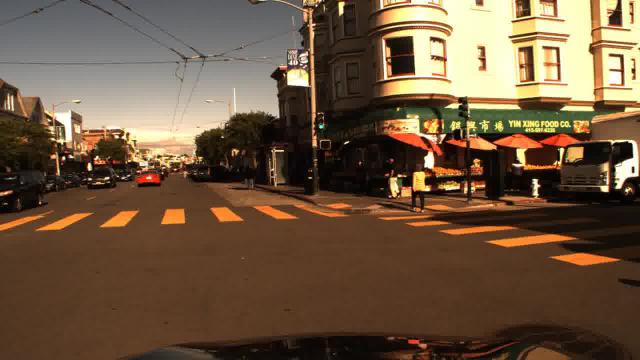}
			\includegraphics[width=.19\linewidth]{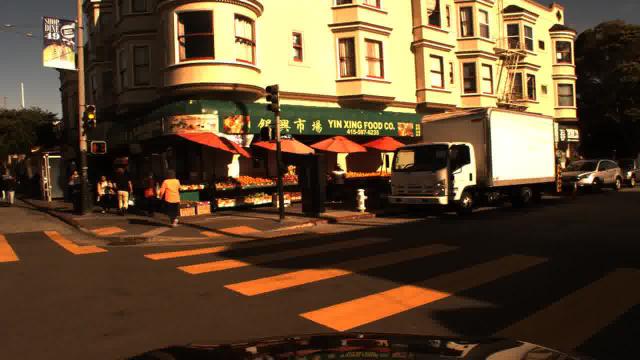}
			\includegraphics[width=.19\linewidth]{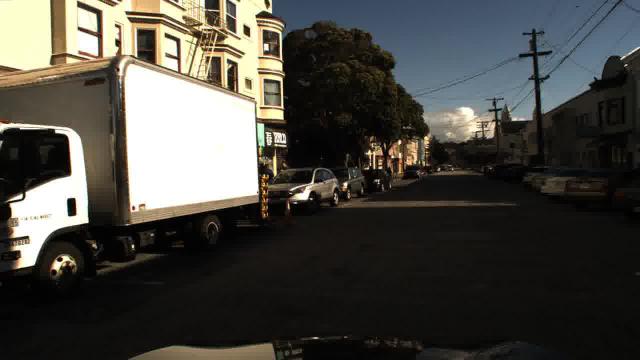}} \\
	(Query) Right Turn with very high uncertainty
	\end{subfigure}
	\begin{subfigure}{1.0\textwidth}
		\centering
		\setlength{\fboxsep}{0pt}%
		\setlength{\fboxrule}{2pt}%
		\fcolorbox{green}{white}{\includegraphics[width=.19\linewidth]{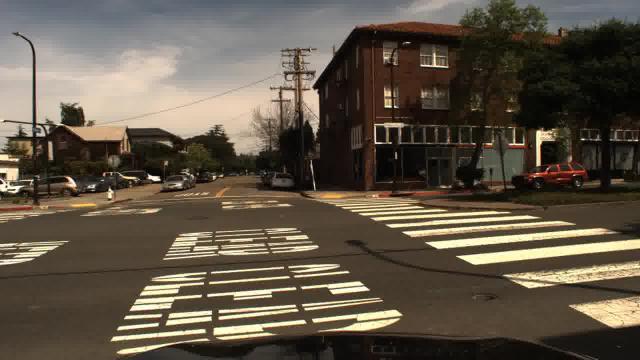}
			\includegraphics[width=.19\linewidth]{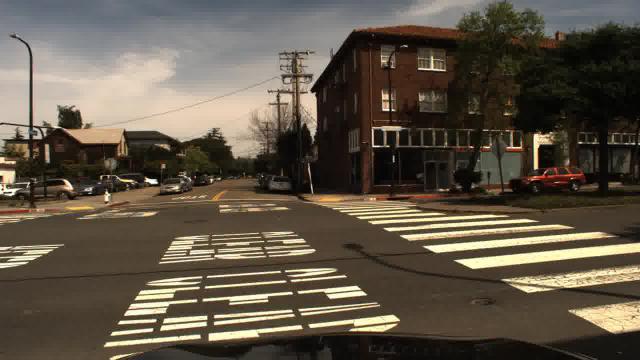}
			\includegraphics[width=.19\linewidth]{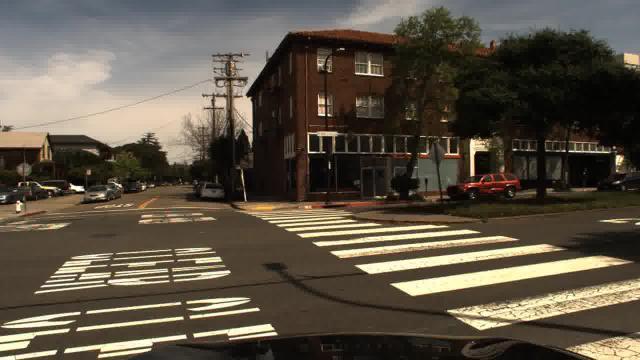}
			\includegraphics[width=.19\linewidth]{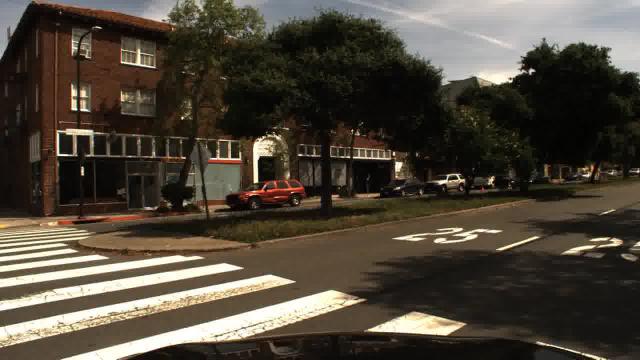}
			\includegraphics[width=.19\linewidth]{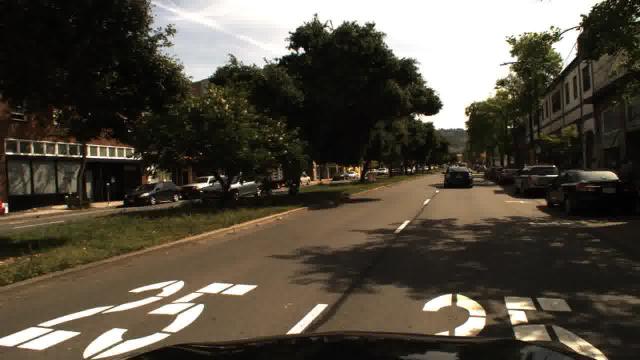}} \\
			(Retrieval result) Right Turn with very low uncertainty 
	\end{subfigure}
	\caption{Qualitative evaluation on HDD using goal-oriented events. Every query is followed by the nearest retrieval result. Outline colors emphasize the event uncertainty degree.  The first query shows a high uncertainty \textit{right lane change}. The second query shows a \textit{right-turn} maneuver blocked by crossing pedestrian. These images are best viewed in color/screen.}
	\label{fig:quality_eval_hdd}
\end{figure*}

Figure~\ref{fig:quality_eval_hdd} presents a qualitative evaluation on HDD. Every two consecutive rows show a very high uncertainty query event and its nearest retrieval result. All query events are chosen within the $1^\text{st}$ highest uncertainty percentile. A description containing the event class and its uncertainty degree is provided below each event. The first query (first row) shows the driver moving from the wrong direction lane to the correct one behind a pickup truck, with a huge cat drawn on a building wall. This example illustrates how uncertainty grounding is challenging in video events. The nearest event to this query (second row) is a very high uncertainty right turn, a similar but not identical event class.

The second query (third row) shows a \textit{right-turn} maneuver where the driver is waiting for crossing pedestrian. The retrieved result (forth row) belongs to the same class but suffers very low uncertainty. We posit the high and low uncertainty are due to pedestrian presence and absence respectively. More visualization using GIFs are available in the supplementary material.

\vspace{-0.05in}
\subsection{Discussion}
We study uncertainty in visual retrieval systems by introducing an extension to the triplet loss. Our unsupervised formulation models embedding space uncertainty. This improves efficiency of the system and identifies confusing visual examples without raising the computational cost. We evaluate our formulation on multiple domains through three datasets. Real-world noisy datasets highlight the utility of our formulation. Qualitative results emphasize our ability in identifying confusing scenarios. This enables data cleaning and reduces error propagation in safety-critical systems.

One limitation of the proposed formulation is bias against minority classes. It treats minority class training samples as noisy input and attenuates their contribution. Tables~\ref{tbl:goal_lbl} and~\ref{tbl:stimuli_lbl} emphasize this phenomenon where performance on the majority and minority classes increases and decreases respectively. Accordingly, micro mAP increases while macro mAP decreases. Thus, this formulation is inadequate for boosting minority classes' performance in imbalanced datasets.

For image applications, high uncertainty is relatively easy to understand -- a favorable quality. Occlusion, inter-class similarity, and multiple distinct instances contribute to visual uncertainty. Unfortunately, this is not the case in video applications, where its challenging to explain uncertainty in events with multiple independent agents. Attention model~\cite{xu2015show,zhou2016learning} is one potential extension, which can ground uncertainty in video datasets.

\section{Conclusion}
We propose an unsupervised ranking loss extension to model local noise in embedding space. Our formulation supports various embedding architectures and ranking losses. It quantifies data uncertainty in visual retrieval systems without raising their computational complexity. This raises stability and efficiency for clean and inherently noisy real-world datasets respectively. Qualitative evaluations highlight our approach efficiency identifying confusing visuals. This is a remarkable add-on for safety-critical domains like autonomous navigation. 

\bibliography{hetero}
\bibliographystyle{icml2019}

\clearpage

\end{document}